\documentclass[10pt,twocolumn,letterpaper]{article}
\usepackage{booktabs}
\usepackage{multirow}
\usepackage{tabularx}
\usepackage{graphicx} % for adjusting table to page width
\usepackage{lipsum} % for dummy text
\usepackage{amssymb} % for checkmark symbols
\usepackage{pifont}  % for additional checkmark symbols
\usepackage{xcolor}
\usepackage{afterpage}
\usepackage{colortbl} % Add this line in the preamble for cell coloring
\usepackage{xcolor} % Add this line in the preamble for color definition
\usepackage{colortbl}
\usepackage{array} 
\usepackage[table]{xcolor} % 使用 xcolor 包来支持表格中的颜色
\usepackage{xcolor}
    \definecolor{Gray}{gray}{0.85}
    \definecolor{LightCyan}{rgb}{0.88,1,1}
\usepackage{tabularray}
    \UseTblrLibrary{booktabs}

\usepackage[pagenumbers]{iccv} % To force page numbers, e.g. for an arXiv version

\definecolor{iccvblue}{rgb}{0.21,0.49,0.74}
\usepackage[pagebackref,breaklinks,colorlinks,allcolors=iccvblue]{hyperref}
   
\def\paperID{6908} % *** Enter the Paper ID here
\def\confName{ICCV}
\def\confYear{2025}

\title{Hints of Prompt: Enhancing Visual Representation for Multimodal LLMs in Autonomous Driving}

\author{
Hao Zhou$^{1,2*}$,
Zhanning Gao$^{2}$, 
Zhili Chen$^{3}$, 
Maosheng Ye$^{2}$, 
Qifeng Chen$^{3}$, 
Tongyi Cao$^{2}$, 
Honggang Qi$^{1\dag}$ \\
$^1$University of Chinese Academy of Sciences, Beijing, China\\
$^2$DeepRoute.AI, Shenzhen, China\\
$^3$The Hong Kong University of Science and Technology, Hong Kong, China\\
{\tt\small zhouhao233@mails.ucas.ac.cn}, 
{\tt\small \{zhanninggao, maoshengye, tongyicao\}@deeproute.ai},\\
{\tt\small zchenei@connect.ust.hk}, 
{\tt\small cqf@cse.ust.hk}, 
{\tt\small hgqi@ucas.ac.cn}
}

\begin{document}
\twocolumn[{
\renewcommand\twocolumn[1][]{#1}
\maketitle

\begin{center}
    \vspace{-16pt}
    \includegraphics[width=0.85\linewidth]{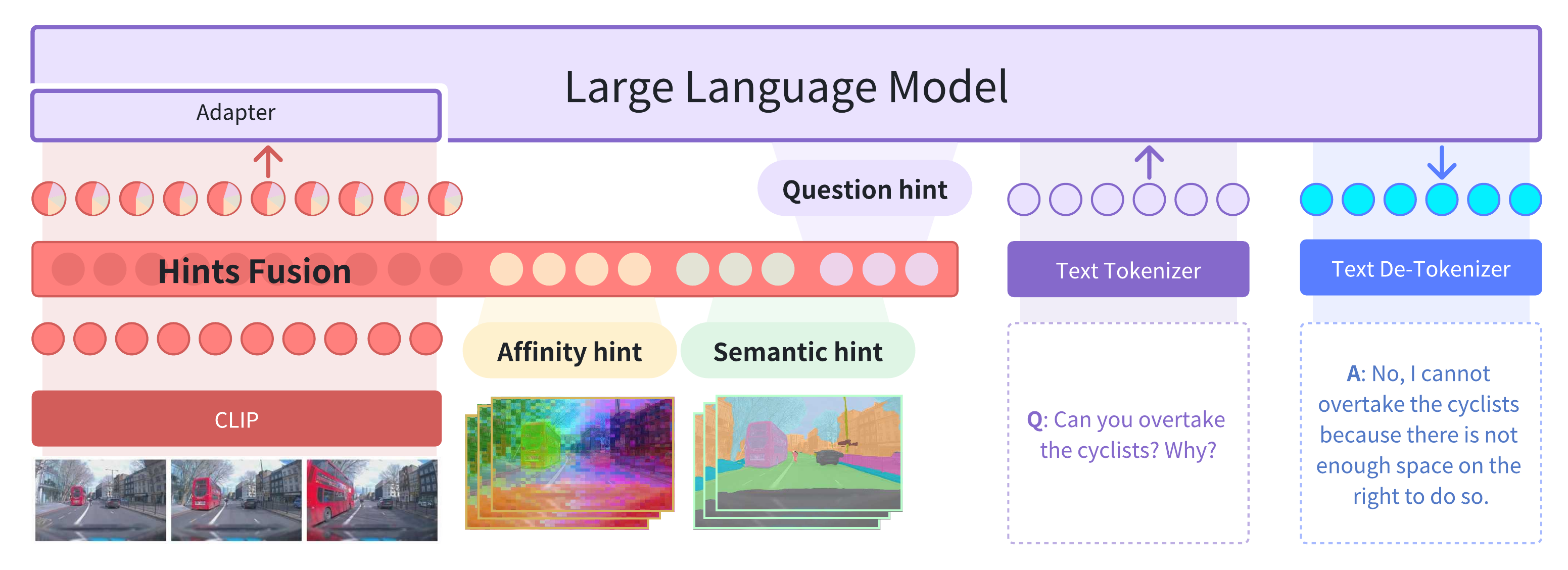}
    \vspace{-16pt}
    \captionof{figure}[Overview of the Hints of Prompt (HoP) Framework]{\textbf{Overview of the Hints of Prompt (HoP) Framework.} HoP enhances Visual Question Answering (VQA) in autonomous driving by incorporating three hierarchical hints: Affinity, Semantic, and Question. The Affinity hint provides foundational instance-level structures through token-wise connections, aiding in instance boundary and interaction recognition. Building on this, the Semantic hint introduces specific instances along with their category information, adding essential driving-related contexts, such as vehicles and traffic signs. Finally, the Question hint guides the LLM’s attention toward image regions pertinent to the question. These hints are fused with visual tokens through a simple Hint Fusion module, aligned via an adapter, and then processed by the LLM to generate answers.}
    \label{fig:framework}
    % \vspace{-2pt}
\end{center}
}]

\renewcommand{\thefootnote}{\fnsymbol{footnote}}
\footnotetext{* Work conducted as intern at DeepRoute.AI. $\dagger$ Corresponding author.}
\renewcommand{\thefootnote}{\arabic{footnote}}

\begin{abstract}
In light of the dynamic nature of autonomous driving environments and stringent safety requirements, general MLLMs combined with CLIP alone often struggle to accurately represent driving-specific scenarios, particularly in complex interactions and long-tail cases. To address this, we propose the Hints of Prompt (HoP) framework, which introduces three key enhancements: Affinity hint to emphasize instance-level structure by strengthening token-wise connections, Semantic hint to incorporate high-level information relevant to driving-specific cases, such as complex interactions among vehicles and traffic signs, and Question hint to align visual features with the query context, focusing on question-relevant regions. These hints are fused through a Hint Fusion module, enriching visual representations by capturing driving-related representations with limited domain data, ensuring faster adaptation to driving scenarios. Extensive experiments confirm the effectiveness of the HoP framework, showing that it significantly outperforms previous state-of-the-art methods in all key metrics.
\end{abstract}

\section{Introduction}
\label{sec:intro}

%In recent years, significant advancements have been made in autonomous driving across academia and industry \cite{ad_survey}\textcolor{red}{[Add more]}. Most industry solutions follow a modular design, including perception \textcolor{red}{[]}, prediction \textcolor{red}{[]}, and planning \textcolor{red}{[]}\cite{ad_system_survey}. While this design simplifies complex problems and enhances interpretability, its cascading structure often leads to information loss and error accumulation throughout the pipeline, which makes it difficult to tackle intricate and infrequent scenarios.

%To address these challenges, 

In recent years, there have been growing trends towards formalizing autonomous driving within end-to-end learning-based frameworks \cite{dl2ad_survey}. These approaches use extensive human driving data to demonstrate human-like behavior and generalization capabilities, enabling them to handle complex scenarios more effectively. However, the end-to-end pipeline often functions as a \textit{black box}, making it challenging to interpret or explain the planning processes driven by neural networks \cite{e2e_survey}. This opacity raises significant ethical and legal concerns regarding accountability and trustworthiness in safety-critical applications.

\begin{figure}[!t]
    \centering
\includegraphics[width=0.8\linewidth]{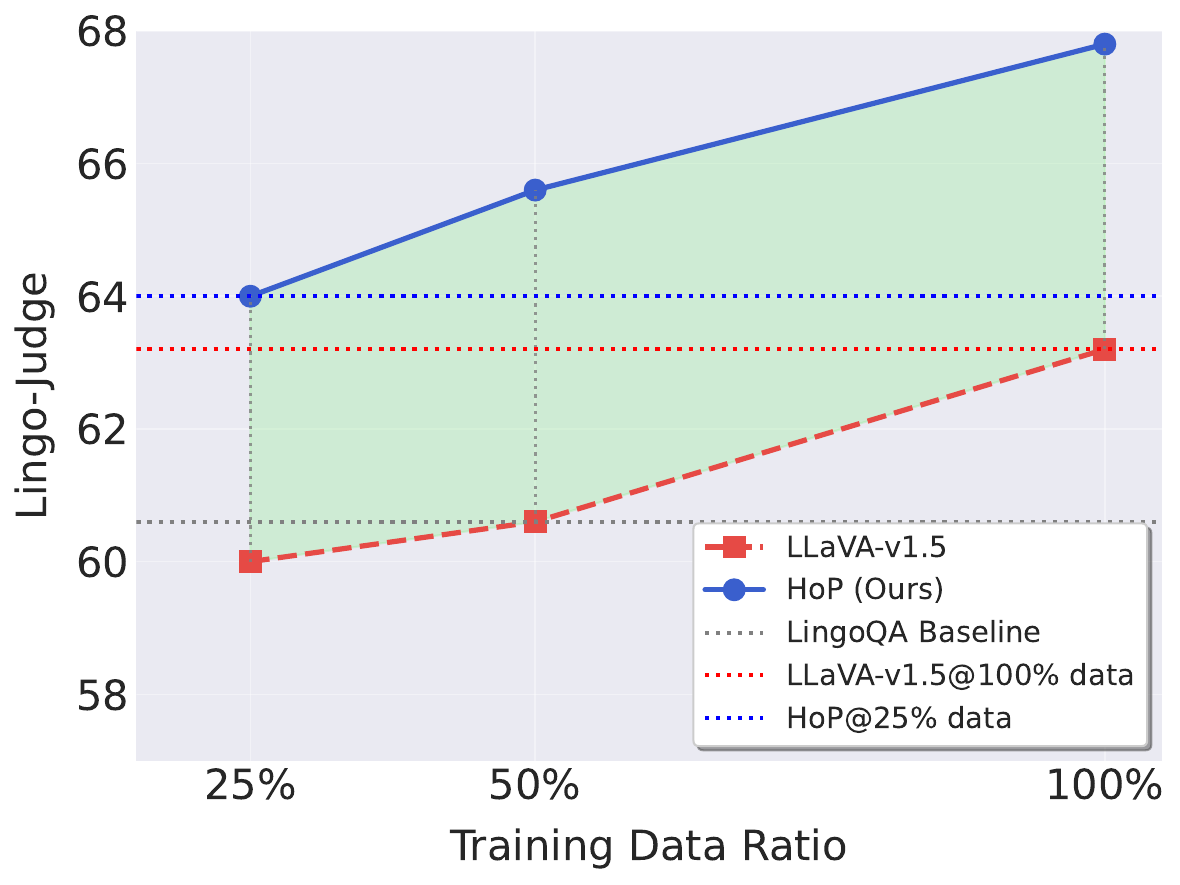}
\vspace{-10pt}
    \caption{Performance under different training data ratio on the LingoQA dataset. HoP surpasses the full-data performance of LLaVA-v1.5 using only \textbf{25\%} of the training data.}
    \label{fig:data_scale}
\end{figure}

Interpretability has become a crucial research focus in autonomous driving to address these concerns. Visual Question Answering (VQA) tasks have gained considerable attention as they allow systems to articulate their observations and actions through natural language interactions \cite{vqa_survey}. 
%By bridging the gap between complex decision-making processes and user comprehension, VQA enhances system transparency and fosters user trust.
The application of Multimodal Large Language Models (MLLMs) \cite{qwen_vl, cogvlm, cambrian, internvl, llava-v1.5, instructblip} has significantly advanced VQA tasks by integrating visual encoders with Large Language Models (LLMs) \cite{palm, llama, vicuna2023}. Typically, MLLMs comprise a visual encoder, an adapter, and an LLM. For VQA, the visual encoder extracts image features, aligned by the adapter with text tokens. The LLM tokenizes the question and integrates visual tokens to generate an answer.

%MLLMs are increasingly applied in autonomous driving to tackle key challenges across perception \cite{drivevlm}, planning \cite{gptdriver, drivevlm}, control \cite{drivegpt4}, question answering (QA) \cite{drivegpt4}, and world modeling \cite{drivedreamer, gaia}. For instance, GPT-Driver \cite{gptdriver} reframes motion planning as a language modeling problem, enhancing decision-making processes, while DriveGPT4 \cite{drivegpt4} predicts control signals and provides real-time explanations. In generative tasks, models like DriveDreamer \cite{drivedreamer} and GAIA-1 \cite{gaia} simulate realistic driving scenarios, supporting data augmentation and facilitating robust training.

%VQA in autonomous driving requires capturing hierarchical visual features essential for complex interactions and long-tail scenarios, while general MLLMs \cite{qwen_vl, cogvlm, llava-v1.5} face the challenge of insufficient domain-specific visual representation due to the limited number of QA pairs in the driving \cite{drivelm, nuscenesqa}. Although existing work \cite{eyes, mome} enhances visual representation through the multi-feature fusion of visual encoders like CLIP \cite{clip}, DINOv2 \cite{dinov2}, and Pix2Struct \cite{pix2struct}, their complex fusion strategies and specific purpose designs fail to meet autonomous driving needs: domain-specific adaptability with limited data.

VQA in autonomous driving requires capturing hierarchical visual features for complex interactions and long-tail scenarios. However, general MLLMs \cite{qwen_vl, cogvlm, llava-v1.5} struggle to capture domain-specific representations due to limited visual-language aligned data in the driving domain \cite{drivelm, nuscenesqa}. Although existing methods \cite{eyes, mome} attempt to enhance visual representation through multi-feature fusion of visual encoders like CLIP \cite{clip}, DINOv2 \cite{dinov2}, and Pix2Struct \cite{pix2struct}, their intricate fusion strategies and specialized designs fall short of addressing the unique demands of autonomous driving: \textit{domain-specific adaptability with limited data}.

%These persistent challenges stem from intrinsic limitations in current multi-encoder fusion methods. First, as shown in \cref{fig:affinity_img_tokens}, CLIP visual tokens lack token-wise affinity relationships critical for preserving instance-level structural integrity, a capability in which DINOv2 \cite{dinov2} excels, but is not utilized. Second, these general visual encoders underrepresent domain-specific semantics, making them prone to missing small and sparse but crucial elements for driving, such as distant vehicles, pedestrians, and traffic signs. Lastly, current MLLMs struggle to adjust visual representations based on specific questions, as visual and text tokens are processed separately. This separation limits the model’s ability to focus on query-relevant image regions, reducing the effectiveness of context-aware responses in autonomous driving.

To enable faster domain adaptation, the visual encoder must quickly capture instance-level information relevant to driving scenarios. However, as shown in \cref{fig:affinity_img_tokens}, CLIP’s visual tokens lack token-wise affinity relationships, which are critical for preserving instance-level structure. Additionally, these general visual encoders underrepresent domain-specific semantics, making them prone to missing small, sparse, yet crucial elements for driving, such as distant vehicles, pedestrians, and traffic signs. Lastly, current MLLMs struggle to adjust visual representations based on specific questions because visual and text tokens are processed separately. This separation limits the model’s ability to focus on query-relevant image regions, reducing the effectiveness of context-aware responses in autonomous driving.

%Based on these observations, we introduce extra tokens as hints to enhance visual representation under limited training data. Specifically, we propose three key enhancements: \textit{Affinity hint} to capture instance-level structures by leveraging token-wise affinity relationships from DINOv2 \cite{dinov2}, \textit{Semantic hint} to introduce sparse driving-specific tokens to incorporate high-level semantic information by using queries from DETR-like models and \textit{Question hint} to correlate visual features according to the query context, ensuring focus on question-relevant regions. We conduct comprehensive studies on designing effective and efficient hints fusion strategies for combining these hint tokens. \cref{fig:framework} illustrates our proposed method, referred to as the Hints of Prompt (HoP) framework throughout the paper. \cref{fig:data_scale} shows HoP can beat the baseline only with \textbf{25\%} training data. Moreover, after analyzing the gain of HoP, we also propose an efficient version of HoP which directly distilled hint tokens from CLIP backbone, efficiently achieving competitive performance.

Based on these observations, we introduce additional tokens as hints to enhance visual representation. Specifically, we propose three key enhancements: \textit{Affinity hint} to capture instance-level structures by leveraging token-wise affinity relationships from DINOv2 \cite{dinov2}; \textit{Semantic hint} to introduce sparse, driving-specific tokens that incorporate high-level semantic information using queries from DETR-like models; and \textit{Question hint} to correlate visual features according to the query context, ensuring focus on question-relevant regions. We conduct comprehensive studies to design effective and efficient fusion strategies for combining these hint tokens. \cref{fig:framework} illustrates our proposed method, referred to as the Hints of Prompt (HoP) framework. \cref{fig:data_scale} shows that HoP outperforms the baseline with only \textbf{25\%} of the training data, demonstrating efficient domain adaptation with limited data in driving scenarios. Furthermore, after analyzing the gains from HoP, we propose an efficient version that directly distills hint tokens from the CLIP backbone, achieving competitive performance more efficiently.

Through extensive experiments on autonomous driving VQA benchmarks, we demonstrate that our method significantly outperforms existing approaches, achieving state-of-the-art results. Our ablation studies further confirm the effectiveness of each hints, as well as the impact of different fusion strategies and LLM scales, illustrating how each component contributes to the overall performance improvement and faster domain adaptation.

To summarize, our contributions are as follows:
\begin{itemize}[leftmargin=1em]
    %\item We identify key limitations in current MLLMs combine with CLIP alone for VQA tasks in autonomous driving and propose an online strategy to enhance visual representations hierarchically.
    % \item We identify key limitations of current general MLLMs with CLIP alone in autonomous driving: insufficient visual understanding with limited data, and domain-agnostic semantic representation.
    \item We identify key limitations of current general MLLMs with CLIP alone in autonomous driving: lack of domain priors for domain-specific visual representation.
    % \item We introduce the Hints of Prompt (HoP) framework, which dynamically enriches visual tokens through a fusion module, significantly enhancing the model’s understanding of complex driving scenes and boosting overall performance.
    \item We introduce the Hints of Prompt (HoP) framework, a data-efficient solution for enhancing visual representation tailored for autonomous driving scenes.
    
    \item Extensive experiments demonstrate significant improvements of our methods, establishing state-of-the-art results on LingoQA \cite{lingoqa}, DRAMA \cite{drama}, and BDD-X \cite{bddx}.
    % \item We consistently observe significant boosts in performance across the datasets of LingoQA\cite{lingoqa}, DRAMA\cite{drama}, and BDD-X\cite{bddx}, establishing a new state-of-the-art for VQA in autonomous driving.
\end{itemize}
\section{Related Work}
\label{sec:re_work}

\textbf{Multimodal Large Language Models.} Multimodal Large Language Models (MLLMs) \cite{llava, llava-v1.5, oryx, internvl, minigpt, gpt4v} have made significant advancements by integrating Large Language Models (LLMs) \cite{vicuna2023, gpt4, llama} with visual encoders, addressing diverse multimodal tasks across various domains. 

In autonomous driving, MLLMs address critical challenges in perception, planning, control, question-answering (QA), and generative world models. For example, GPT-Driver \cite{gptdriver} reframes motion planning as a language modeling task, enhancing accuracy, while DriveGPT4 \cite{drivegpt4} combines vision and language input to predict control signals with real-time explanations, improving transparency. 

In perception, MLLMs advance sensor data fusion from LiDAR and cameras; models such as Talk2BEV \cite{talk2bev} use Bird’s Eye View (BEV) features for complex visual tasks, while LMDrive \cite{lmdrive} and DriveVLM \cite{drivevlm} employ trajectory-based reasoning to refine scene understanding and decision making. In QA, DriveLM \cite{drivelm} uses graph-based reasoning to enhance real-time planning. For world models, generative approaches like DriveDreamer \cite{drivedreamer} and GAIA-1 \cite{gaia} generate realistic driving scenarios for data augmentation and training. These advancements highlight MLLMs’ potential to enhance multimodal understanding and optimize autonomous driving systems across diverse applications.\vspace{0.5\baselineskip}

\noindent
\textbf{Enhancing Visual Representations in MLLMs.} Effective alignment of visual representations with LLM capabilities is essential to improve overall performance. Taking advantage of data, InternVL \cite{internvl} introduces a large-scale visual encoder to improve granularity of the features, and some works \cite{deepseek_vl, cambrian, eyes, mome, sphinx} combine multiple visual encoders with a complex fusion strategy to take advantage of complementary strengths. However, effective alignment of visual representations with LLM capabilities remains challenging in autonomous driving. 

QA-ViT \cite{qa_vit} incorporates textual information into visual features to help MLLMs focus on relevant image regions, enhancing interpretability while maintaining efficiency. Similarly, we utilize the same LLM text embeddings as Question hints, but our fusion strategy is more easily integrated with various visual backbones. While Hint-AD \cite{hint-ad} aligns language with perception-planning tokens holistically, and TOKEN \cite{token} leverages enriched representations generated by an end-to-end driving model to mitigate data scarcity, our method uniquely introduces dynamic hint tokens with a lightweight fusion module, enabling context-aware visual-semantic alignment for driving scenarios.

\section{Method}
\label{sec:method}
\subsection{Overview}

The Hints of Prompt (HoP) framework (\cref{fig:framework}) leverages three distinct token types—Affinity, Semantic, and Question hints—each representing a different level of information. These hints progress from instance-level, token-wise connections in the Affinity hint, to high-level semantic information in the Semantic hint, and finally to question-specific context in the Question hint, gradually refining and aligning the visual representation with the LLM. A simple Hint Fusion module integrates these hints into a cohesive visual representation, which is then processed by an adapter and passed to the LLM for answer generation. By incorporating hints across different information levels, our method enhances context-aware in VQA for autonomous driving and facilitates domain adaptation under data scarcity.

\begin{figure}[!t]
    \centering
\includegraphics[width=0.8\linewidth]{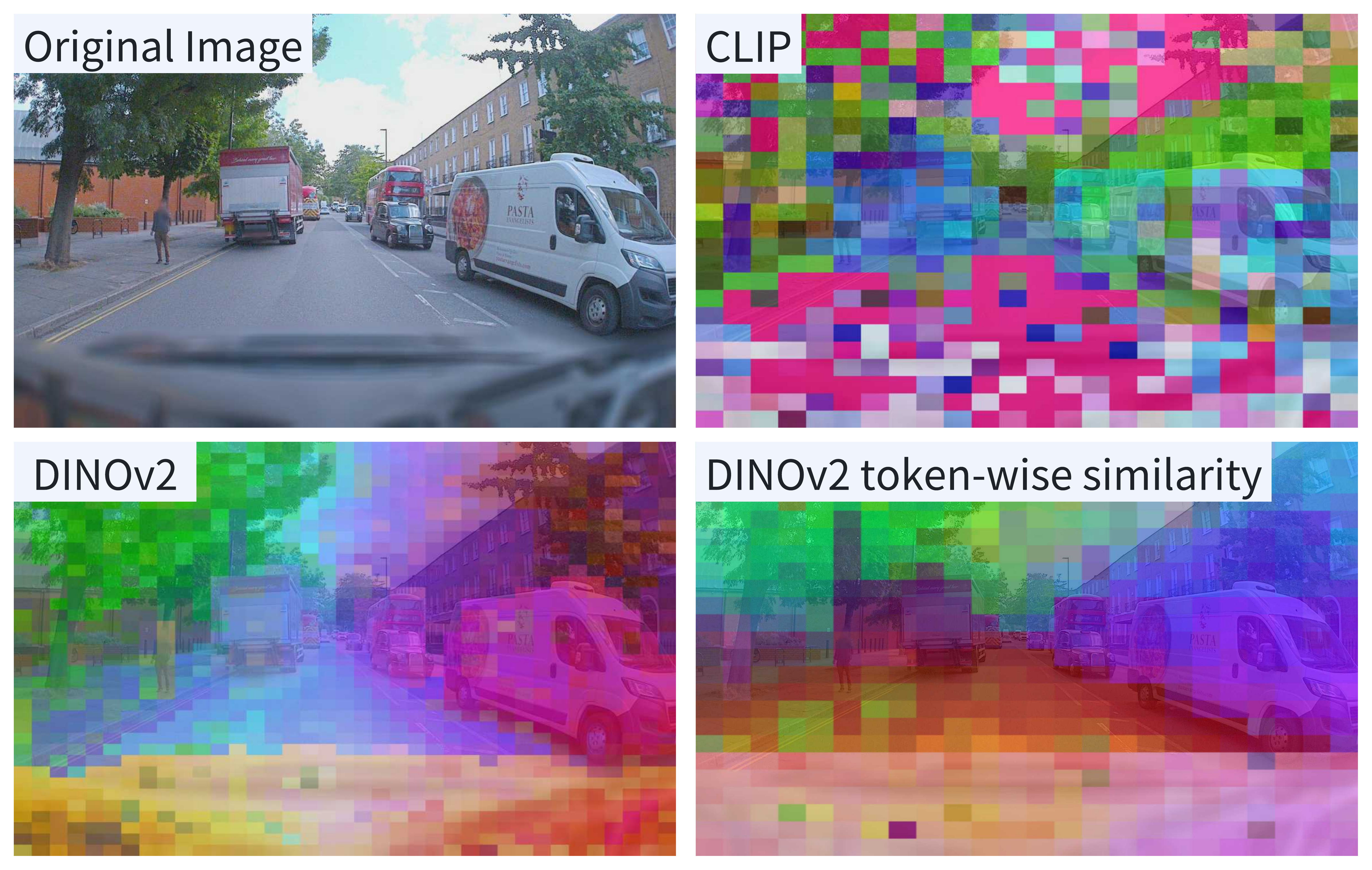}
    \caption{Visualization of token affinity from CLIP \cite{clip} and DINOv2 \cite{dinov2}. Similar colors indicate higher affinity scores, with color values derived from a PCA-reduced token vector space. DINOv2 token-wise similarity denotes tokens embedded only from the similarity matrix.}
    \label{fig:affinity_img_tokens}
\end{figure}

\begin{figure*}[!t]
    \centering
    \includegraphics[width=0.9\linewidth]{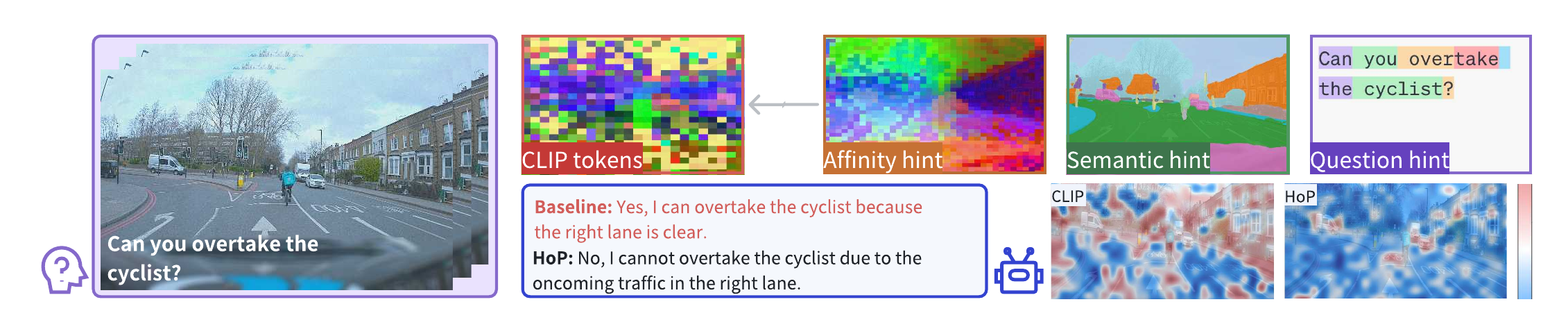}
    \vspace{-0.4cm}
    \caption{The impact of the three proposed hint types compared to the baseline and attention maps with/without these hints.}
    \label{fig:case_study}
\end{figure*}
\subsection{Hint Tokens}
\label{subsec:hinttokens}

\textbf{Affinity hint} The default practice of MLLMs \cite{llava, deepseek_vl, llava-v1.5} applies the CLIP \cite{clip} encoder for visual feature extraction, where the features are well-aligned with the text embedding space and contain rich semantic knowledge. However, the CLIP visual tokens struggle to understand driving scenes that require spatial reasoning among instances on the road. As shown in \cref{fig:affinity_img_tokens}, we observe that the CLIP visual tokens drop the affinity information among instances, which might hinder the following LLM from reasoning upon these limited visual tokens. We find that DINOv2 \cite{dinov2} maintains significant affinity among tokens within the same instances, offering a more coherent spatial representation. Therefore, we use DINOv2 tokens as Affinity hint tokens to reinforce the instance-level structure in complex driving scenes.

To verify that the enhancement is due to token-wise affinity rather than specific DINOv2 features, we experiment with similarity information among the DINOv2 visual tokens and observe a notable performance gain. A detailed discussion is provided in \cref{subsec:ablation_study}.

\vspace{0.2\baselineskip}
\noindent
\textbf{Semantic hint} After implementing instance-level enhancement, providing a semantic category specific to the driving domain can further focus visual representations on key instances, such as the cyclist shown in \cref{fig:case_study}. The Semantic hint aims to incorporate high-level, domain-specific semantic information, such as vehicles, traffic signs, and pedestrians, which is crucial for understanding complex interactions. To achieve this, we employ sparse queries from DETR-like \cite{detr} detection or segmentation pipelines as Semantic hints. Specifically, GroundingDINO \cite{grounding_dino} or Mask2Former \cite{mask2former} will be used for the extraction of Semantic hints. Experimental results (\cref{tab:semantic_hint_backbone}) show that benefiting from Mask2Former’s adaptation to driving scenarios, it can align more effectively with the task-specific requirements of VQA in autonomous driving.

\vspace{0.2\baselineskip}
\noindent
\textbf{Question hint} In VQA tasks, different questions often shift attention to specific regions of an image. For example, a question about the weather may focus on broader aspects, such as the sky or overall lighting. In contrast, a question about pedestrians in the path directs attention to finer details and specific areas. Incorporating question-specific information can further refine this focus by enhancing visual representation to capture key instances in the driving environment. Aligning the visual representation with each question’s particular demands enables a more targeted, context-aware interpretation, thus enhancing relevance and effectiveness in complex driving scenarios. Similar to QA-ViT \cite{qa_vit}, we found that using LLM's text tokens as the Question hint effectively achieves context-aware representation. We also explored an alternative approach that utilizes CLIP’s text encoder to generate the Question hint. A comparison of these methods is shown in \cref{tab:question_hint_ablation}.

As shown in \cref{fig:case_study}, incorporating the three proposed hints improves the model’s response over the baseline. The baseline incorrectly suggests overtaking is possible, overlooking oncoming traffic. As shown by the HoP attention map, the model with hints gains instance-level structure and focuses more accurately on key elements, such as the cyclist and oncoming vehicles. This added context enhances the model’s spatial reasoning, resulting in a more precise and relevant response in complex driving scenarios. Detailed quantitative analysis is available in \cref{sec:experiments}.

\begin{table*}[ht]
\centering
\begin{booktabs}{
        colsep = 11pt,
        colspec = {lccccc}, % Column format
        row{9-10} = {bg=gray!15}, % Set last rows with gray background
        cell{1}{2-6} = {c}, % Center-align header row
    }
    \toprule
    Method & LLM & \textbf{Lingo-Judge} $\uparrow$ & BLEU-4 $\uparrow$ & METEOR $\uparrow$ & CIDEr $\uparrow$ \\
    \midrule
    GPT-4V \cite{gpt4v} (zero-shot) & -  & 59.6 & 6.30 & 12.4 & 42.8 \\
    LingoQA Baseline \cite{lingoqa} & Vicuna-v1.5-7B & 60.6 & 15.0 & 18.6 & 65.6 \\
    QA-ViT \cite{qa_vit}†         & Vicuna-v1.5-7B & 63.6 & 14.6 & 19.0 & 65.9 \\
    VTS \cite{vts}           & InternLM2-7B & 64.2 & 14.5 & \textbf{20.5} & 56.9 \\
    LLaVA-v1.5 \cite{llava-v1.5}† & Vicuna-v1.5-7B & 63.2 & 14.1 & 19.3 & 63.7 \\
    LLaVA-v1.5 (DINOv2)† & Vicuna-v1.5-7B & 62.8 & 14.7 & 19.3 & \underline{68.3} \\
    LLaVA-v1.5 (+A-MoF) \cite{eyes}† & Vicuna-v1.5-7B & 64.2 & 14.5 & 19.1 & 
    64.7\\
    \midrule
    Efficient HoP      & Vicuna-v1.5-7B & \underline{66.8} & \underline{15.2} & 20.0 & 66.2 \\
    HoP      & Vicuna-v1.5-7B & \textbf{67.8} & \textbf{15.8} & \underline{20.3} & \textbf{70.9} \\
    \bottomrule
\end{booktabs}
\vspace{-8pt}
\caption{\textbf{Performance comparison on the LingoQA dataset.} The best-performing method for each metric is highlighted in \textbf{bold}, while
the second-best method is indicated by an \underline{underline}. ``†" indicates model that are fine-tuned on the LingoQA dataset. \textit{Efficient HoP} represents the lightweight version of our HoP method, while \textit{LLaVA-v1.5} serves as our baseline. \textit{LLaVA-v1.5 (DINOv2)} replaces the CLIP visual encoder with a DINOv2 encoder. The HoP method consistently demonstrates superior performance across all metrics.}
\label{tab:lingoqa_result}
\end{table*}

\subsection{Hints Fusion}
\begin{figure}[!t]
    \centering
    \includegraphics[width=1.0\linewidth]{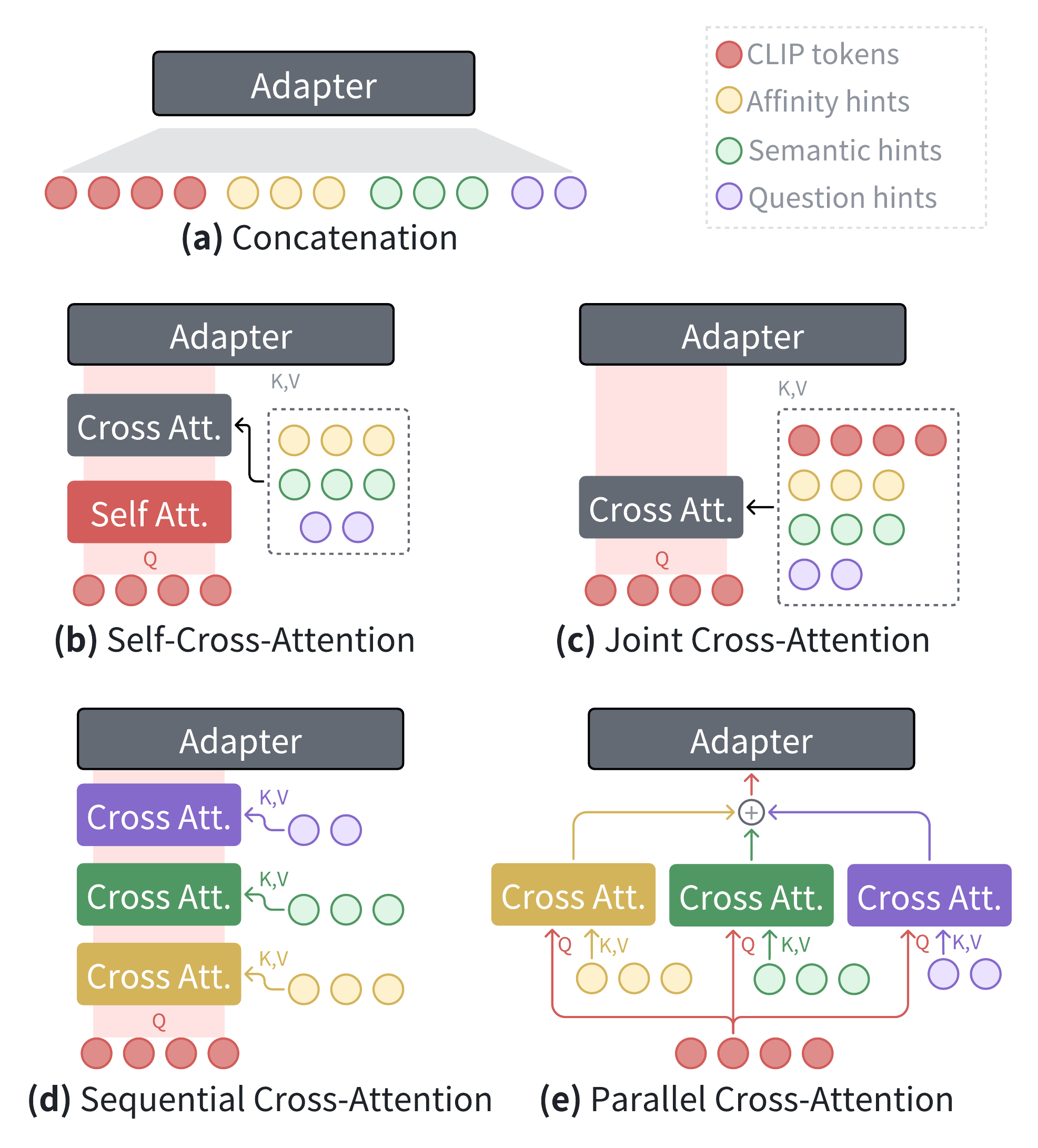}
    \vspace{-0.8cm}
    \caption{Different fusion strategies for the Hints Fusion module. The joint cross-attention strategy demonstrates the best performance. Residual connections are omitted for simplicity.}
    \label{fig:fusion_strategy}
\end{figure}

We explore several fusion strategies to merge the above hints into the original CLIP \cite{clip} visual tokens. Formally, we denote Affinity, Semantic, and Question hints as \(\mathbf{H}_\text{A} \in \mathbb{R}^{N \times d}\), \(\mathbf{H}_\text{S} \in \mathbb{R}^{M \times d}\), and \(\mathbf{H}_\text{Q} \in \mathbb{R}^{K \times d}\), respectively, where \(N\), \(M\), and \(K\) are the number of tokens for each hint, and \(d\) is the embedding dimension. The original CLIP visual tokens are represented as \(\mathbf{P} \in \mathbb{R}^{L \times d}\), and the fused visual tokens are denoted as \(\mathbf{P}_\text{f} \in \mathbb{R}^{L \times d}\). The notation \(\texttt{[·]}\) indicates the concatenation operation, \(\texttt{CA}(\mathbf{P}, \mathbf{H})\) denotes a Cross-Attention layer with \(\mathbf{P}\) as query and \(\mathbf{H}\) as key-value, and \(\texttt{SA(·)}\) denotes a Self-Attention layer. As shown in \cref{fig:fusion_strategy}, we present the different fusion strategies as follows.

\vspace{0.2\baselineskip}
\noindent
\textbf{Concatenation} As a baseline, hint tokens are directly concatenated with CLIP tokens, as illustrated in \cref{fig:fusion_strategy}(a):
\begin{equation}
  \mathbf{P}_\text{f} = \texttt{[}\mathbf{P}, \mathbf{H}_\text{A}, \mathbf{H}_\text{S}, \mathbf{H}_\text{Q}\texttt{]}.
  \label{eq:concat_fusion}
\end{equation}
Interestingly, we find that simple concatenation can harm visual representation and degrade performance (see \cref{tab:hints_fusion}), suggesting that directly injecting multi-level information may confuse the adapter and LLM, hindering relevant information extraction.

% hz: This strategy is simple, intuitive, and easy to implement, making it a natural baseline for exploring more complex fusion mechanisms.
\vspace{0.2\baselineskip}
\noindent
\textbf{Self-Cross-Attention} To enable deeper interaction between visual tokens and hint tokens, we propose a Self-Cross-Attention module (\cref{fig:fusion_strategy}(b)) that mimics the standard Transformer \cite{transformer} layer but omits the feed-forward network (FFN) to accommodate the subsequent adapter module:
\begin{equation}
    \mathbf{P}_\text{f} = \mathbf{P} + \texttt{CA}(\texttt{SA}(\mathbf{P}), \texttt{[}\mathbf{H}_\text{A}, \mathbf{H}_\text{S}, \mathbf{H}_\text{Q}\texttt{]}).
\end{equation}
This allows refinement of the visual tokens before incorporating the hint tokens.

\vspace{0.2\baselineskip}
\noindent
\textbf{Joint Cross-Attention} As shown in \cref{fig:fusion_strategy}(c), inspired by Flamingo \cite{flamingo}, Joint Cross-Attention omits the self-attention stage and uses the visual representation \( \mathbf{P} \) as both query and key-value in a single cross-attention layer:
\begin{equation}
    \mathbf{P}_\text{f} = \mathbf{P} + \texttt{CA}(\mathbf{P},\texttt{[}\mathbf{P}, \mathbf{H}_\text{A}, \mathbf{H}_\text{S}, \mathbf{H}_\text{Q}\texttt{]}).
\end{equation} 
The design simplifies the fusion process without weakening the interaction between visual tokens and hint tokens.

To compare the effects of different query-hint interaction strategies, we further explore two additional approaches: \textit{Parallel Cross-Attention} and \textit{Sequential Cross-Attention}, as shown in \cref{fig:fusion_strategy}(d, e). In these strategies, hint tokens first interact individually with the query and are then fused through either a parallel or cascading way.

\noindent
\textbf{Sequential Cross-Attention} sequentially applies cross-attention between the visual representation and each hint, progressively enriching the representation:
\begin{equation}
    \mathbf{P}_\text{f} = \mathbf{P} + \texttt{CA}(\texttt{CA}(\texttt{CA}(\mathbf{P}, \mathbf{H}_\text{A}), \mathbf{H}_\text{S}), \mathbf{H}_\text{Q}).
\end{equation}

% \vspace{0.5\baselineskip}
\noindent
\textbf{Parallel Cross-Attention} applies cross-attention simultaneously between the visual representation and each hint token, which allows all hint tokens to influence the visual representation in parallel:
\begin{equation}
    \mathbf{P}_\text{f} = \mathbf{P} + \texttt{CA}(\mathbf{P}, \mathbf{H}_\text{A}) + \texttt{CA}(\mathbf{P}, \mathbf{H}_\text{S}) + \texttt{CA}(\mathbf{P}, \mathbf{H}_\text{Q}).
\end{equation}
% \vspace{0.5\baselineskip}

% zhili: 
% We intriguingly found that simple concatenation might harm the visual representation and degrade performance, suggesting that directly injecting information from different levels confuses the adapter and LLM when extracting relevant information. We empirically found that our proposed Joint Cross-Attention achieves the best performance while maintaining efficiency across various VQA tasks in autonomous driving while maintaining a comparable efficiency.

% haozhou:
% add: While this strategy has proven effective, other fusion methods may still offer potential improvements, suggesting that there is room for further exploration.

We empirically find that the proposed Joint Cross-Attention achieves the best performance with high efficiency across diverse VQA tasks in autonomous driving, outperforming the other fusion strategies mentioned above. More details are provided in \cref{subsec:ablation_study}.

%While this approach proves highly effective, alternative fusion methods may offer additional improvements, suggesting room for further exploration.

%Surprisingly, simple concatenation may harm the visual representation and degrade performance, suggesting that directly injecting information at different levels without fusion strategies makes it difficult for the adapter and LLM to extract relevant information, confusing.

\subsection{Efficient Hints of Prompt}

% Although introducing extra hint tokens significantly improves VQA performance, extracting these hints adds computational burden. As mentioned in \cref{subsec:hinttokens} and demonstrated in \cref{tab:affinity_sources}, using a similarity matrix as affinity hint tokens can enhance VQA performance. Based on these observations, to reduce resource overhead, we propose an efficient version of HoP. The main idea is to distill or fine-tune lightweight necks to mimic the behavior of DINOv2 \cite{dinov2} or Mask2former \cite{mask2former}, providing additional instance- and semantic-level hints to enhance the visual representation while maintaining low latency.

Although introducing extra hint tokens significantly improves VQA performance, extracting these hints adds a computational burden. To reduce resource overhead, we propose an efficient version of HoP. The main idea is to distill or fine-tune lightweight heads to mimic the behavior of DINOv2 \cite{dinov2} or Mask2Former \cite{mask2former}, providing additional instance-level and semantic-level hints to enhance the visual representation while maintaining low latency. More details are provided in \cref{subsec:implementation_details} and \cref{subsec:ablation_study}.

\section{Experiments}
\label{sec:experiments}
We evaluate the Hints of Prompt (HoP) across three datasets: LingoQA \cite{lingoqa}, DRAMA \cite{drama}, and BDD-X \cite{bddx}, focusing on its performance in VQA tasks for autonomous driving. Our experiments encompass overall performance improvements, along with ablation studies on the design of each hint, fusion strategies, and efficiency.

% \begin{table*}[ht]
% \centering
% \begin{booktabs}{
%         colsep = 11pt,
%         colspec = {lccccc}, % 列格式
%         row{8-9} = {bg=gray!15}, % 设置最后一行为灰色背景
%         cell{1}{2-6} = {c}, % 标题行居中对齐
%     }
%     \toprule
%     Method & LLM & \textbf{Lingo-Judge} $\uparrow$ & BLEU-4 $\uparrow$ & METEOR $\uparrow$ & CIDEr $\uparrow$ \\
%     \midrule
%     GPT-4V \cite{gpt4v} (zero-shot) & -  & 59.6 & 6.30 & 12.4 & 42.8 \\
%     LingoQA Baseline \cite{lingoqa} & Vicuna-v1.5-7B & 60.6 & 15.0 & 18.6 & 65.6 \\
%     LLaVA-v1.5(DINOv2) & Vicuna-v1.5-7B & 62.8 & 14.7 & 19.3 & 68.3 \\
%     QA-ViT \cite{qa_vit}         & Vicuna-v1.5-7B & 63.6 & 14.6 & 19.0 & 65.9 \\
%     LLaVA-v1.5 & Vicuna-v1.5-7B & 63.2 & 14.1 & 19.3 & 63.7 \\
%     VTS \cite{vts}           & InternLM2-7B & 64.2 & 14.5 & \textbf{20.5} & 56.9 \\
%     \midrule
%     Efficient HoP      & Vicuna-v1.5-7B & 66.8 & 15.2 & 20.0 & 66.2 \\
%     HoP      & Vicuna-v1.5-7B & \textbf{67.8} & \textbf{15.8} & 20.3 & \textbf{70.9} \\
%     \bottomrule
% \end{booktabs}
% \caption{\textbf{Performance comparison on the LingoQA dataset.} \textit{Efficient HoP} denotes the lightweight version of our HoP method, while \textit{LLaVA-v1.5} serves as our baseline. \textit{LLaVA-v1.5 (DINOv2)} replaces the CLIP visual encoder with a DINOv2 encoder. The HoP method consistently demonstrates superior performance across all metrics.}
% \label{tab:lingoqa_result}
% \end{table*}

\begin{table*}[ht]
\centering
% \resizebox{\textwidth}{!}{
\begin{booktabs}{
        colsep = 15pt,
        colspec = {lcccccc}, % 列格式
        row{5-6} = {bg=gray!15}, % 设置最后一行为灰色背景
        cell{1}{2-7} = {c}, % 标题行居中对齐
    }
    \toprule
    Method & BLEU-1$\uparrow$ & BLEU-4$\uparrow$ & METEOR$\uparrow$ & ROUGE$\uparrow$ & CIDEr$\uparrow$ & SPICE$\uparrow$ \\
    \midrule
    LCP \cite{drama} & 73.9 & 54.7 & 39.1 & 70.0 & \textbf{3.7} & 56.0 \\
    VTS \cite{vts} & 75.3 & 55.8 & 40.7 & 74.7 & 2.8 & 58.0 \\
    LLaVA-v1.5† & 75.8 & 56.1 & 41.0 & 78.0 & \underline{2.9} & 58.4 \\
    \midrule
    Efficient HoP & \underline{76.0} & \underline{56.2} & \underline{41.3} & \underline{78.5} & 2.7 & \underline{58.8} \\
    HoP & \textbf{76.2} & \textbf{56.3} & \textbf{41.7} & \textbf{79.8} & 2.8 & \textbf{59.1} \\
    \bottomrule
\end{booktabs}
% }
\vspace{-8pt}
\caption{\textbf{Performance comparison on the DRAMA dataset.} ``†" indicates models that are fine-tuned on the DRAMA dataset. HoP demonstrates enhanced performance over other methods across most metrics, with \textit{LLaVA-v1.5} as the baseline.}
\label{tab:drama_results}
\end{table*}

\begin{table*}[ht]
\centering
\small
\begin{booktabs}{
        colspec = {l|ccc|ccc|ccc|ccc}, % 列格式
        row{6-7} = {bg=gray!15}, % 设置指定行的灰色背景
        cell{1-2}{2-13} = {c}, % 设置标题行居中对齐
        colsep = 3.4pt, % 设置列间距
    }
    \toprule
    \SetCell[r=2]{l} Method & & Easy & & & Medium & & & Hard & &  & All \\
    \cmidrule{2-4}\cmidrule{5-7}\cmidrule{8-10}\cmidrule{11-13}
    & CIDEr$\uparrow$ & B4$\uparrow$ & ROUGE$\uparrow$ & CIDEr$\uparrow$ & B4$\uparrow$ & ROUGE$\uparrow$ & CIDEr$\uparrow$ & B4$\uparrow$ & ROUGE$\uparrow$ & CIDEr$\uparrow$ & B4$\uparrow$ & ROUGE$\uparrow$ \\
    \midrule
    ADAPT \cite{adapt} & 100.93 & \textbf{20.90} & 46.17 & 62.66 & 16.44 & \underline{40.80} & 52.71 & \textbf{13.56} & 40.49 & 85.38 & 17.40 & 43.04 \\
    DriveGPT4 \cite{drivegpt4} & 113.20 & \underline{20.38} & 46.46 & \underline{65.01} & 16.94 & 40.51 & 57.29 & 12.28 & 42.07 & 99.10 & \textbf{18.32} & 44.73 \\
    LLaVA-v1.5† & 111.84 & 19.83 & 47.57 & 63.76 & 14.69 & 40.34 & 49.80 & 11.60 & 41.97 & 96.51 & 17.42 & 45.42 \\
    \midrule
    Efficient HoP & \underline{118.30} & 19.95 & \textbf{47.65} & 64.80 & \underline{17.58} & 40.73 & \underline{58.20} & 12.92 & \underline{42.58} & \underline{99.80} & 17.70 & \underline{45.65} \\
    HoP & \textbf{121.01} & 20.09 & \underline{47.60} & \textbf{65.16} & \textbf{18.02} & \textbf{40.81} & \textbf{60.37} & \underline{13.01} & \textbf{43.52} & \textbf{102.20} & \underline{17.91} & \textbf{45.79} \\
    \bottomrule
\end{booktabs}
\vspace{-8pt}
\caption{\textbf{Performance comparison across different splits of the BDD-X testing dataset.} “†” indicates model that are fine-tuned on the BDD-X dataset. The evaluation follows the DriveGPT4 \cite{drivegpt4} protocol, reporting CIDEr, BLEU-4 (B4), and ROUGE scores across Easy, Medium, Hard, and Overall (All) splits. \textit{LLaVA-v1.5} serves as the baseline.}
\label{tab:bddx_results}
\end{table*}

% \begin{table}[h]
% \centering
% \begin{tabular}{lccc}
% \toprule
% \multirow{2}{*}{Dataset Ratio} & \multicolumn{3}{c}{Lingo-Judge$\uparrow$} \\
% \cmidrule(lr){2-4}
% & LLaVA-v1.5 & HoP (Ours) & $\Delta$ \\
% \midrule
% 25\% & 60.0 & {64.0} & \textcolor{green!70!black}{+4.0} \\
% 50\% & 60.6 & 65.6 & \textcolor{green!70!black}{+5.0} \\
% 100\% & 63.2 & 67.8 & \textcolor{green!70!black}{+4.6} \\
% \bottomrule
% \end{tabular}
% \caption{\textbf{Data-efficient domain adaptation of HoP.} $\Delta$: the performance gain of HoP over LLaVA-v1.5 at same data ratio.}
% \label{tab:lingoqa_data_efficiency}
% \end{table}

\subsection{Implementation Details}
\label{subsec:implementation_details}
We implement the Hints of Prompt (HoP) method based on LLaVA-v1.5 \cite{llava-v1.5}, freezing the visual encoder and fully fine-tuning the remaining modules.

For the Affinity hint, we use DINOv2-large \cite{dinov2} to extract 576 hint tokens by downsampling the feature map to align with the CLIP tokens ($24 \times 24$). To verify that the improvement is due to token-wise affinity rather than specific DINOv2 features, we compute a similarity matrix from the 576 DINOv2 tokens and use it as hint tokens, processed through a linear layer. Results are shown in \cref{tab:affinity_sources}.

For the Semantic hint, we select the top-K queries from Mask2Former \cite{mask2former} or GroundingDINO \cite{grounding_dino} as hint tokens, based on confidence scores, with each token adding an embedding feature that corresponds to the label index. The Question hint uses the LLM’s textual embeddings associated with question tokens.

To create Efficient HoP, we distill and fine-tune two lightweight heads from the CLIP backbone: a 4-layer, 8-heads, 512-dimensional ViT-like \cite{vit} decoder for the Affinity hint, distilled via cosine similarity to DINOv2 features on the LingoQA training dataset; and a ViTDet-like \cite{vitdet} neck with a Mask2Former head for the Semantic hint, trained on the Cityscapes dataset \cite{Cityscapes}. The training setup follows that of the original Mask2Former.

The whole HoP training is conducted for one epoch using the AdamW optimizer with a cosine learning rate schedule, a warm-up ratio of 0.03, and a base learning rate of 2e-5. All models are trained on 32 NVIDIA A100 GPUs with a batch size of 4 per GPU. We use an NVIDIA RTX 4090 GPU to measure latency in the inference.

\subsection{Datasets and Benchmarks}
\textbf{LingoQA} \cite{lingoqa} is a benchmark dataset for video question answering in autonomous driving, with 28K unique driving scenarios and 419K question-answer pairs. The dataset consists of two subsets: \textit{Action} and \textit{Scenery}, plus a high-quality \textit{Evaluation} subset for benchmarking. The main metric, Lingo-Judge, is an evaluation model aligned with human preferences, while BLEU \cite{papineni2002bleu}, METEOR~\cite{banerjee2005meteor}, and CIDEr~\cite{vedantam2015cider} provide lexical similarity assessments.

\vspace{0.3\baselineskip}
\noindent
\textbf{DRAMA} \cite{drama} dataset comprises 17K video clips of interactive urban traffic scenes in Tokyo, and provides extensive annotations for autonomous driving risk assessment. It includes lots of VQA pairs about 11.8K training samples and 2.5K testing samples. Evaluation metrics include BLEU, METEOR, CIDEr, ROUGE \cite{lin2004rouge}, and SPICE \cite{anderson2016spice}.

\vspace{0.3\baselineskip}
\noindent
\textbf{BDD-X} \cite{bddx} is a dataset developed for autonomous driving tasks, focusing on video captioning and rationales, we adopt the VQA pairs and evaluation protocol provided by DriveGPT4 \cite{drivegpt4}, which include 26K training samples and 1.8K testing samples. BLEU, CIDEr, and ROUGE are employed as metrics.

% \begin{table}[!t]
% \centering
% \setlength{\tabcolsep}{3pt}
% \begin{tabular}{cccccccc}
% \toprule
% ID & AH & SH & QH & LJ $\uparrow$ & B4 $\uparrow$ & METEOR $\uparrow$ & CIDEr $\uparrow$ \\
% \midrule
% 0 &  &   &  & 63.2 & 14.1 & 19.3 & 63.7 \\
% \midrule
% 1 & \makebox[1.5em]{\checkmark} &  &  & 64.6 & 14.6 & 19.0 & 66.0 \\
% 2 &  & \makebox[1.5em]{\checkmark} &  & 64.6 & 14.4 & 19.2 & 67.8 \\
% 3 &  &  & \makebox[1.5em]{\checkmark} & 63.6 & 14.3 & 18.9 & 66.3 \\
% \midrule
% 4 & \makebox[1.5em]{\checkmark} & \makebox[1.5em]{\checkmark} &  & 65.4 & 15.2 & 20.2 & 66.8 \\
% 5 & \makebox[1.5em]{\checkmark} &  & \makebox[1.5em]{\checkmark} & 65.2 & 14.9 & 19.3 & 68.3 \\
% 6 &  & \makebox[1.5em]{\checkmark} & \makebox[1.5em]{\checkmark} & 65.8 & 15.2 & 19.7 & 68.4 \\
% \midrule
% 7 & \makebox[1.5em]{\checkmark} & \makebox[1.5em]{\checkmark} & \makebox[1.5em]{\checkmark} & \multicolumn{1}{c}{67.2} & \multicolumn{1}{c}{15.3} & \multicolumn{1}{c}{20.1} & \multicolumn{1}{c}{69.2} \\
% 8 & \makebox[1.5em]{\checkmark} & \makebox[1.5em]{\checkmark{\textsubscript{+cls}}} & \makebox[1.5em]{\checkmark} & \multicolumn{1}{c}{\textbf{67.8}} & \multicolumn{1}{c}{\textbf{15.8}} & \multicolumn{1}{c}{\textbf{20.3}} & \multicolumn{1}{c}{\textbf{70.9}} \\
% \bottomrule
% \end{tabular}
% \caption{\textbf{The impact of each hint type within the HoP framework.} Abbreviations: AH (Affinity hint), SH (Semantic hint), QH (Question hint), LJ (Lingo-Judge), and B4 (BLEU-4). “+cls” denotes the addition of class information to the Semantic hint.}
% \label{tab:ablation-hints}
% \end{table}

\begin{table}[!t]
\centering
\setlength{\tabcolsep}{4.5pt} % Adjust column spacing for compactness
\begin{tabular}{lcccc}
\toprule
AH types & LJ $\uparrow$ & B4 $\uparrow$ & METEOR $\uparrow$ & CIDEr $\uparrow$ \\
\midrule
Original tokens & \textbf{64.6} & 14.6 & 19.0 & 66.0 \\
Similarity tokens & 64.2 & \textbf{15.5} & \textbf{19.3} & \textbf{71.1} \\
\bottomrule
\end{tabular}
\vspace{-8pt}
\caption{\textbf{Impact of different Affinity hint types on performance.} This ablation examines only the Affinity hint without incorporating other hint types. “Original tokens” refer to DINOv2 tokens, while “Similarity tokens” are derived from the similarity matrix. Abbreviations: AH (Affinity Hint), LJ (Lingo-Judge), B4 (BLEU-4).}
\label{tab:affinity_sources}
\end{table}

\subsection{Comparison with Existing Works}
Our results on the LingoQA \cite{lingoqa}, DRAMA \cite{drama}, and BDD-X \cite{bddx} datasets demonstrate the effectiveness of the Hints of Prompt (HoP) method in enhancing VQA for autonomous driving. As shown in \cref{tab:lingoqa_result}, HoP significantly outperforms baseline on LingoQA, achieving a top Lingo-Judge score of \textbf{67.8} and leading BLEU-4, METEOR, and CIDEr scores.

As shown in \cref{tab:drama_results}, HoP surpasses other methods on the DRAMA dataset across key metrics, including BLEU-1, BLEU-4, METEOR, ROUGE, and SPICE, highlighting its capability to generate detailed and contextually accurate answers in urban traffic scenes.

In \cref{tab:bddx_results}, HoP’s performance on BDD-X, evaluated using the DriveGPT4 protocol, consistently outperforms other methods across all difficulty levels (Easy, Medium, Hard, and All), achieving the highest CIDEr scores and robust results even in challenging driving scenarios. 

In addition, the Efficient HoP, a lightweight version of our method, also performs competitively in all datasets, demonstrating the scalability of our approach.

\begin{figure*}[!t]
    \centering
    \includegraphics[width=0.80\linewidth]{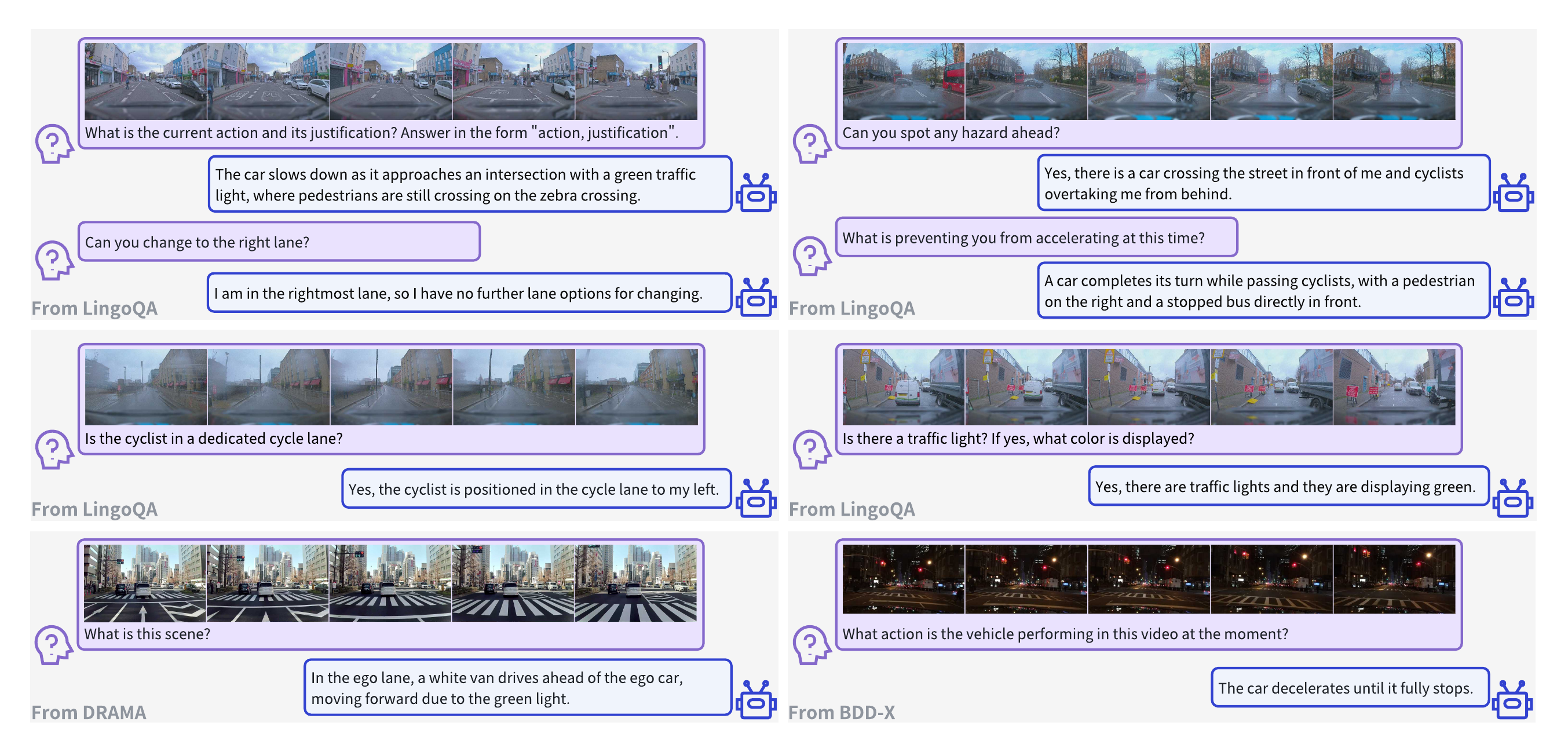}
    \vspace{-0.65cm}
    \caption{\textbf{Visualization results across datasets.} Example responses generated by our method on the LingoQA \cite{lingoqa}, DRAMA \cite{drama}, and BDD-X \cite{bddx} datasets. }
    \label{fig:vis}
\end{figure*}

Visualization examples can be found in \cref{fig:vis}.

% \begin{table}[!t]
% \centering
% \setlength{\tabcolsep}{4pt}
% \begin{tabular}{lcccc}
% \toprule
% +AH types & LJ $\uparrow$ & B4 $\uparrow$ & METEOR $\uparrow$ & CIDEr $\uparrow$ \\
% \midrule
% Original tokens & \textbf{64.6} & 14.6 & 19.0 & 66.0 \\
% Similarity tokens & 64.2 & \textbf{15.5} & \textbf{19.3} & \textbf{71.1} \\
% \bottomrule
% \end{tabular}
% \caption{\textbf{Comparison of Affinity Hint sources on LingoQA.} Performance comparison between DINOv2 feature and DINOv2 token-wise similarity as sources for Affinity Hint.}
% \label{tab:affinity_sources}
% \end{table}

\begin{table}[!t]
\centering
\setlength{\tabcolsep}{3pt}
\begin{tabular}{lcccc}
\toprule
 SH types & LJ $\uparrow$ & B4 $\uparrow$ & METEOR $\uparrow$ & CIDEr $\uparrow$ \\ 
\midrule
GroundingDINO \cite{grounding_dino} & 66.4 & 15.2 & 20.1 & 68.4 \\
Mask2Former \cite{mask2former} & \textbf{67.8} & \textbf{15.8} & \textbf{20.3} & \textbf{70.9} \\
\bottomrule
\end{tabular}
\vspace{-8pt}
\caption{\textbf{Impact of different Semantic hint types}. Abbreviations: SH (Semantic hint), LJ (Lingo-Judge), B4 (BLEU-4).}
\label{tab:semantic_hint_backbone}
\end{table}

\begin{table}[!t]
\centering
\setlength{\tabcolsep}{7pt}
\begin{tabular}{lcccc}
\toprule
QH types & LJ $\uparrow$ & B4 $\uparrow$ & METEOR $\uparrow$ & CIDEr $\uparrow$ \\ 
\midrule
CLIP & 65.2 & 15.3 & 20.1 & 67.0 \\
LLM & \textbf{67.8} & \textbf{15.8} & \textbf{20.3} & \textbf{70.9} \\
\bottomrule
\end{tabular}
\vspace{-8pt}
\caption{\textbf{Impact of different Question hint types.} “CLIP” refers to hint tokens derived from the CLIP text encoder, “LLM” refers to tokens from the LLM input embedding layer. Abbreviations: QH (Question Hint), LJ (Lingo-Judge), B4 (BLEU-4).}
\label{tab:question_hint_ablation}
\end{table}

\subsection{Ablation Study}
\label{subsec:ablation_study}

\begin{figure}[!t]
    \centering
    \includegraphics[width=0.9\linewidth]{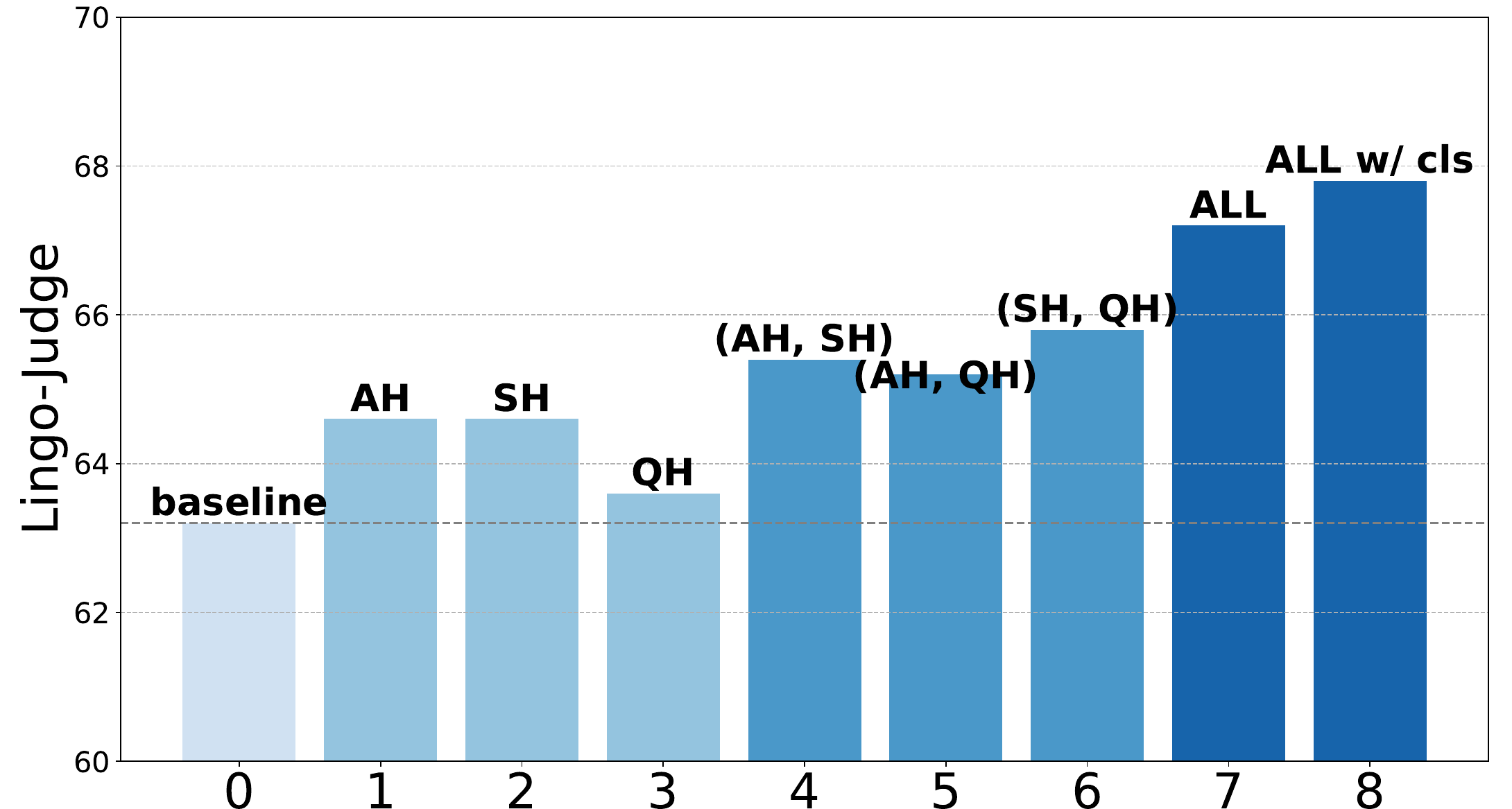}
    \vspace{-0.2cm}
    \caption{\textbf{Impact of each hint type within the HoP.} Abbreviations: AH (Affinity hint), SH (Semantic hint), QH (Question hint). “w/ cls” indicates the addition of class information to the Semantic hint, “ALL” denotes the combination of three hint types.}
    \label{fig:ablation-hints}
\end{figure}

\begin{figure}[!t]
    \centering
    \includegraphics[width=0.8\linewidth]{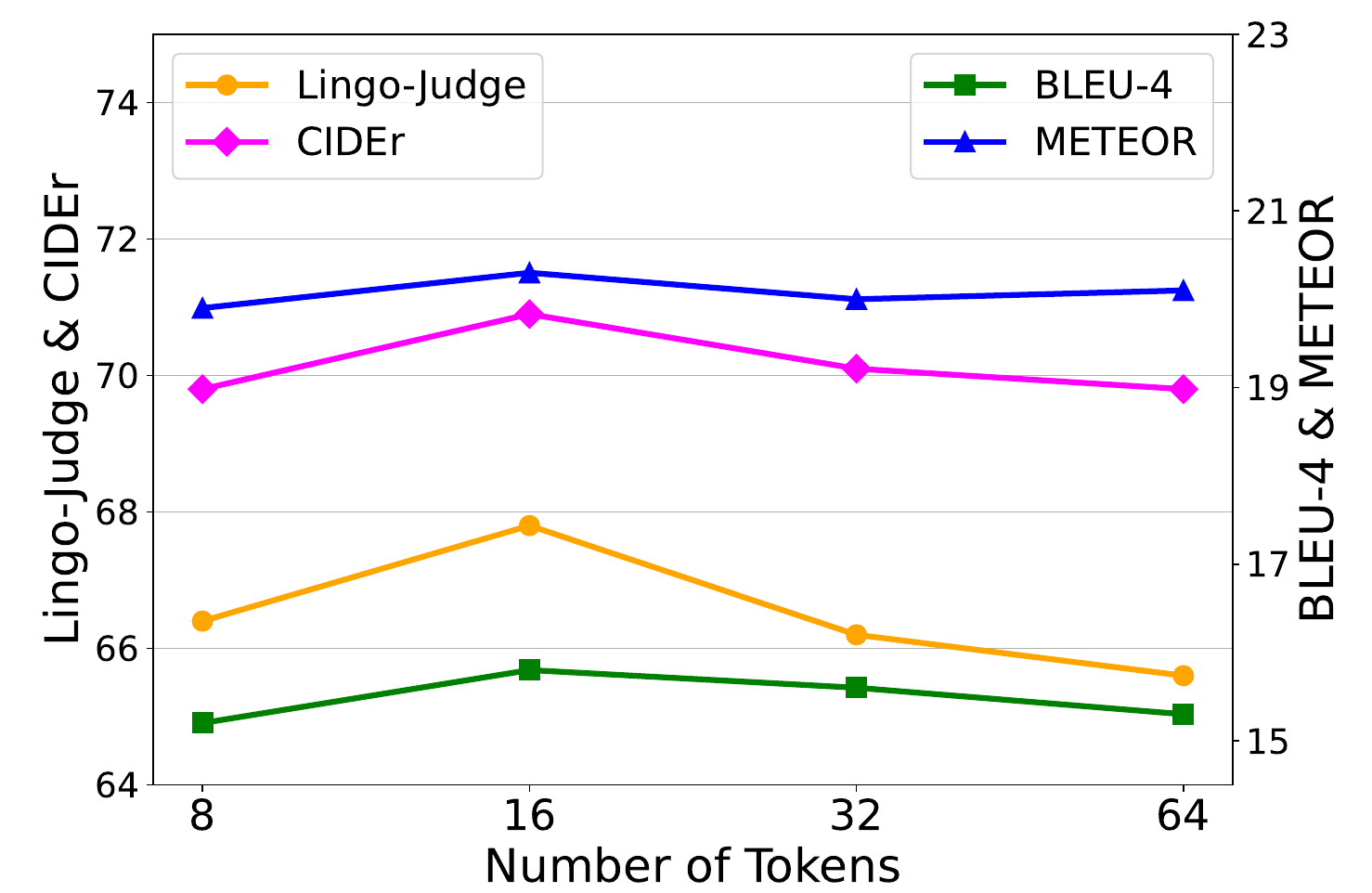}
    \vspace{-0.2cm}
    \caption{The effect of the number of Semantic hint tokens on performance in the LingoQA \cite{lingoqa} dataset.}
    \label{fig:num_of_tokens}
\end{figure}

To evaluate the contributions of each component within the HoP framework, we conduct comprehensive ablation studies. These include data-efficient domain adaptation of HoP (\cref{fig:data_scale}), analyzing the effectiveness of each hint type independently (\cref{fig:ablation-hints}), assessing the impact of different hint sources (\cref{tab:affinity_sources}, \cref{tab:semantic_hint_backbone}, \cref{tab:question_hint_ablation}), and hyperparameters (\cref{fig:num_of_tokens}). We also evaluate various fusion strategies for integrating the hints (\cref{tab:hints_fusion}) and, finally, analyze the efficiency gains of the proposed Efficient HoP (\cref{tab:efficient_hop}).

\vspace{0.3\baselineskip}
\noindent
\textbf{Data-efficient domain adaptation of HoP} In \cref{fig:data_scale}, HoP trained with only 25\% data achieves 64.0 Lingo-Judge score, surpassing the full-data performance of LLaVA-v1.5 (63.2), demonstrates remarkable data efficiency in driving-specific VQA task. More details are in the appendix.

\vspace{0.3\baselineskip}
\noindent
\textbf{Combination of Different Hint Types} \cref{fig:ablation-hints} shows an ablation study on the LingoQA dataset, assessing the impact of each hint type within the HoP framework. Applying each hint independently (IDs 1–3) consistently improves performance over the baseline (ID 0) across all metrics, highlighting each hint’s unique contribution to model comprehension. Combining two hints (IDs 4–6) provides additional gains, while using all three hints together (IDs 7–8) achieves the highest scores overall, demonstrating their complementary effects. Notably, adding class information to the Semantic hint (+cls in ID 8) further enhances performance.

\vspace{0.3\baselineskip}
\noindent
\textbf{Effectiveness of Affinity Hint} As shown in Table \ref{tab:affinity_sources}, using DINOv2 \cite{dinov2} token-wise similarity as the source for the Affinity hint provides notable improvements over the baseline, indicating that the core benefit of the Affinity hint lies in capturing token-wise similarity. Although using the full DINOv2 features as the Affinity hint achieves slightly better results, this can be attributed to the additional feature information present in DINOv2. These results confirm the effectiveness of the Affinity hint and underscore the importance of token-wise connections in representing driving scenarios.

\vspace{0.3\baselineskip}
\noindent
\textbf{Impact of Different Semantic Hint} As shown in \cref{tab:semantic_hint_backbone}, Mask2Former \cite{mask2former} achieves superior performance across all metrics, with a Lingo-Judge score of 67.8 compared to 66.4 for GroundingDINO \cite{grounding_dino}. This boost suggests that Mask2Former’s domain adaptation through training on Cityscapes \cite{Cityscapes} allows it to capture driving-specific semantic details more effectively. In contrast, while GroundingDINO excels in open-set object detection, its lack of domain-specific training limits its alignment with the task-specific requirements of VQA in autonomous driving.

% \begin{table}[ht]
% \centering
% \small
% \resizebox{\linewidth}{!}{
% \begin{tabular}{lcccc}
% \toprule
% Number of Queries & LJ $\uparrow$ & BLEU-4 $\uparrow$ & METEOR $\uparrow$ & CIDEr $\uparrow$ \\ 
% \midrule
% 8   & 66.4 & 15.2 & 19.9 & 69.8 \\ 
% 16  &  \textbf{67.8} & \textbf{15.8} & \textbf{20.3} & \textbf{70.9} \\
% 32  & 66.2 & 15.6 & 20.0 & 70.1 \\ 
% 64  & 65.6 & 15.3 & 20.1 & 69.8 \\ 
% \bottomrule
% \end{tabular}}
% \caption{Impact of the number of object queries on POP model performance. Selecting 16 object queries achieves the best performance while increasing the number to 64 leads to a performance drop due to noise from irrelevant or erroneous detections.}
% \label{tab:object_query_analysis}
% \end{table}

We also analyze the impact of varying the number of Semantic hint tokens on performance in the LingoQA dataset. As shown in \cref{fig:num_of_tokens}, increasing the token count from 8 to 16 improves performance across all metrics. Adding more tokens up to 32 yields similar results, while increasing the count to 64 leads to a decline, likely due to noise introduced by less relevant queries from Mask2Former.

\vspace{0.3\baselineskip}
\noindent
\textbf{Impact of Different Question Hint Sources} We evaluate two sources for extracting question embeddings for the Question hint: CLIP’s text model and the LLM embedding layer. As shown in \cref{tab:question_hint_ablation}, using the LLM embedding layer yields higher performance across all metrics, achieving a Lingo-Judge score of 67.8 compared to 65.2 with CLIP’s text model. This result suggests that the LLM embeddings offer superior alignment for multimodal reasoning tasks.

\begin{table}[!t]
\centering
\setlength{\tabcolsep}{2.5pt}
\resizebox{\columnwidth}{!}{
\begin{tabular}{lccc}
\toprule
Fusion Strategy & LJ $\uparrow$ & \#Params (M)  & GFLOPs  \\ 
\midrule
Concatenation & 58.8 & - & - \\ 
Self-Cross-Attention & 65.2 & \underline{12.9} & \underline{5.0} \\ 
Parallel Cross-Attention & 66.6 & 17.1 & 8.63 \\ 
Sequential Cross-Attention & \underline{67.0} & 17.1 & 8.63 \\ 
Joint Cross-Attention & \textbf{67.8} & \textbf{8.7}& \textbf{3.8} \\  
\bottomrule
\end{tabular}
}
\vspace{-8pt}
\caption{\textbf{Impact of different fusion strategies for integrating hints.} Joint Cross-Attention delivers the best performance with efficient parameter and computational use.}
\label{tab:hints_fusion}
\end{table}

% \begin{table}[!t]
% \centering
% \setlength{\tabcolsep}{12pt}
% \begin{tabular}{lcc}
% \toprule
% Method & \#Params (M) & Latency (ms) \\
% \midrule
% LLaVA-v1.5  & 303.5 & 631 \\
% VTS  & 304.0 & 740 \\
% \midrule
% HoP & 832.0 & 956 \\
% Efficient HoP  & \textbf{385.8} &\textbf{661}\\
% \bottomrule
% \end{tabular}
% \caption{\textbf{Efficiency of the Efficient HoP.} \#Params indicate the parameter count for all modules related to \textit{visual representation}. Latency reports the average inference time measured on the LingoQA \cite{lingoqa} validation set on an NVIDIA RTX 4090 GPU.}
% \label{tab:eff_hop_vs_hop}
% \end{table}

% \begin{table}[h]
%     \centering
%     \small
%     \setlength{\abovecaptionskip}{1pt}
%     \resizebox{\columnwidth}{!}{
%     \begin{tabular}{l c c c}
%         \toprule
%         Method & LLM & LJ $\uparrow$ & Latency (ms) \\
%         \midrule
%         \multirow{3}{*}{Efficient HoP} 
%           & Qwen-v2-0.5B & 57.6$_{+2.8}$ & 302 \\
%           & Qwen-v2.5-3B & \textbf{64.6$_{+3.4}$} & 504 \\
%           & Qwen-v2.5-3B (AWQ \cite{awq}) & 64.2$_{+3.0}$ & \textbf{281} \\
%         \bottomrule
%     \end{tabular}
%     }
%     \caption{Performance and latency of Efficient HoP variants. Subscripts denote improvements over the baseline (without HoP). AWQ represents the result after quantizing the model.}
%     \label{tab:efficient_hop}
% \end{table}

\begin{table}[!t]
    \centering
    \setlength{\tabcolsep}{2.5pt}
    \resizebox{\columnwidth}{!}{
    \begin{tabular}{l c c c c}
    \toprule
    Method & LLM & LJ $\uparrow$ & \#Params (M) & Latency (ms) \\
    \midrule
    VTS  &InternLM2-7B & 64.2 & 304.0 & 740 \\
    LLaVA-v1.5  &Vicuna-v1.5-7B & 63.2 & 303.5 & 631 \\
    HoP &Vicuna-v1.5-7B & \textbf{67.8} & 832.0 & 956 \\
    \midrule
    \multirow{3}{*}{Efficient HoP} 
      & Vicuna-v1.5-7B & 66.8 & 385.8 & 661\\
      & Qwen-v2.5-3B & 64.6 & 385.8 & 504 \\
      & Qwen-v2-0.5B & 57.6 & 385.8 & 302 \\
    with AWQ \cite{awq} &Qwen-v2.5-3B & 64.2 & 385.8 & \textbf{281} \\
        \bottomrule
    \end{tabular}
    }
    \vspace{-8pt}
    \caption{\textbf{Performance and efficiency of Efficient HoP.} \#Params indicate the parameter count for all modules related to \textit{visual representation}. Latency reports the average inference time measured on the LingoQA \cite{lingoqa} validation set on an NVIDIA RTX 4090 GPU. AWQ \cite{awq} represents the quantized model.}
    \label{tab:efficient_hop}
\end{table}

\vspace{0.3\baselineskip}
\noindent
\textbf{Hint Fusion Strategy} As shown in \cref{tab:hints_fusion}, the Joint Cross-Attention strategy delivers the best performance, achieving the highest Lingo-Judge (LJ) score of 67.8. This method is not only effective but also highly efficient, requiring only 8.7M parameters and 3.8 GFLOPs—significantly less than the Self-Cross-Attention. In comparison, Sequential and Parallel Cross-Attention strategies, while competitive with LJ scores of 67.0 and 66.6, respectively, require nearly twice the computational resources of Joint Cross-Attention.

\vspace{0.3\baselineskip}
\noindent
\textbf{Efficient HoP} In the original HoP framework, we use DINOv2 \cite{dinov2} and Mask2Former \cite{mask2former} for Affinity and Semantic hints, which boost performance but increase computational costs. To reduce these, we create an Efficient HoP. As shown in \cref{tab:efficient_hop}, Efficient HoP reduces parameters and latency compared with HoP while achieving comparable performance (\cref{tab:lingoqa_result}, \cref{tab:drama_results}, and \cref{tab:bddx_results}). The quantized Efficient HoP with the 3B model maintains this advantage with deployment-ready 281ms latency.
\section{Conclusion}
\label{sec:conclusion}
In this paper, we introduce the Hints of Prompt (HoP) framework to enhance the visual representation of MLLM in VQA tasks for autonomous driving scenarios. By incorporating Affinity, Semantic, and Question hints, HoP effectively captures instance-level structure, domain-specific elements, and query-specific context, enhancing interpretability and reasoning capabilities, and enabling rapid domain adaptation with limited data. Extensive experiments on the LingoQA, DRAMA, and BDD-X datasets demonstrate that HoP consistently achieves state-of-the-art performance. Moreover, we propose an Efficient HoP variant that reduces computational overhead while maintaining high performance. We believe that HoP’s structured integration of multi-level hints offers a promising direction for advancing VQA in complex, safety-critical scenarios.

\clearpage

\section*{Acknowledgement}
This work was supported by the National Natural Science Foundation of China under Grant No. 62271466 and the National Key Research and Development Program of China Grant No. 2023YFF0716500.
{
    \small
    \bibliographystyle{ieeenat_fullname}
    \bibliography{main}
}


\end{document}